\documentclass[acmsmall]{acmart}

\setcopyright{cc}
\setcctype{by}
\acmJournal{PACMMOD}
\acmYear{2025} \acmVolume{3} \acmNumber{6 (SIGMOD)} \acmArticle{357} \acmMonth{12} \acmPrice{}\acmDOI{10.1145/3769822}


\received{April 2025}
\received[revised]{July 2025}
\received[accepted]{August 2025}

\usepackage{multirow}
\usepackage{todonotes}
\usepackage{listings}
\usepackage{soul}
\usepackage{xcolor}
\usepackage[dvipsnames]{xcolor}
\usepackage[most]{tcolorbox}
\usepackage{algorithm}
\usepackage{algorithmic}
\usepackage{textcomp}
\RequirePackage[inline,shortlabels]{enumitem}
\usepackage{longtable}
\usepackage{booktabs}    
\usepackage{multirow}

\newcommand{\sysname}{TeCoD}

\newcommand{\vx}{\mathbf{x}}
\newcommand{\vy}{\mathbf{y}}
\newcommand{\ty}{\tilde{\vy}}
\newcommand{\tx}{\tilde{\vx}}
\newcommand{\vg}{\mathbf{g}}
\newcommand{\textx}{\mathbf{q}}
\newcommand{\testx}{\textx}
\newcommand{\matched}{{\tt Matched}}
\newcommand{\unmatched}{{\tt Unmatched}}
\newcommand{\tset}{\mathcal{T}}
\newcommand{\indexT}{$\mathcal{I}$}

\newcommand{\SCD}{Fixed Template}
\newcommand{\GCD}{Flexible Template}
\newcommand{\hlred}[1]{\textcolor{red}{#1}}

\begin{document}
\title{Reliable Answers for Recurring Questions: Boosting Text-to-SQL Accuracy with Template Constrained Decoding}

\author{Smit Jivani}
\email{smitjivani@iitb.ac.in}
\orcid{0009-0003-1743-055X}
\author{Sarvam Maheshwari}
\email{sarvam@iitb.ac.in}
\orcid{0009-0005-4133-6357}
\author{Sunita Sarawagi}
\email{sunita@iitb.ac.in}
\orcid{0009-0003-1743-055X}
\affiliation{%
  \institution{Department of Computer Science and Engineering, Indian Institute of Technology Bombay}
  \city{Mumbai}
  \country{India}
}

\pagebreak

\begin{abstract}

Large language models (LLMs) have revolutionised Text-to-SQL generation, allowing users to query structured data using natural language with growing ease. Yet, real-world deployment remains challenging, especially in complex or unseen schemas, due to inconsistent accuracy and the risk of generating invalid SQL. 

We introduce Template Constrained Decoding (\sysname), a system that addresses these limitations by harnessing the recurrence of query patterns in labeled workloads. \sysname\ converts historical NL-SQL pairs into reusable templates and introduces a robust template selection module that uses a fine-tuned natural language inference model to match or reject queries efficiently. Once the template is selected, TeCoD enforces it during SQL generation through grammar-constrained decoding, implemented via a novel partitioned strategy that ensures both syntactic validity and efficiency. Together, these components yield up to 36\% higher execution accuracy than in-context learning (ICL) and 2.2× lower latency on matched queries.

\end{abstract}


\maketitle

%



\section{Introduction}
\label{sec:intro}
\begin{figure*}[h]
    \centering
    \includegraphics[width=\textwidth]{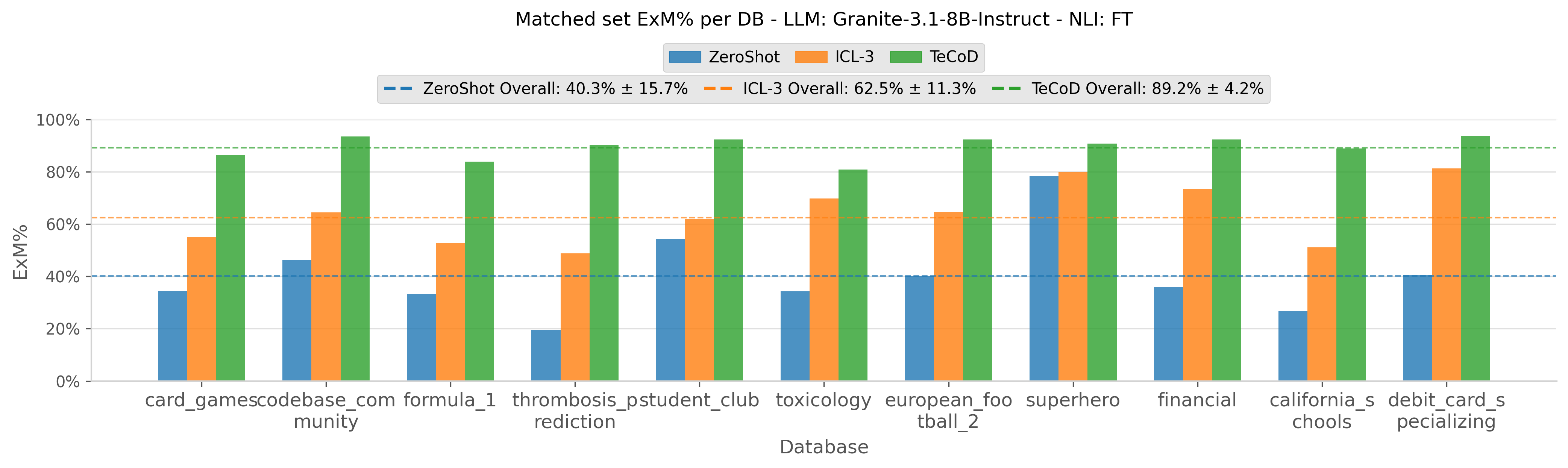}
    \caption{Execution Match Accuracy(\%) per database for recurring questions (questions with a matching template). Observe that \sysname\ provides close to 90\% accuracy across almost all databases, whereas both ZeroShot and ICL are much worse and exhibit high variance.}
    \label{fig:zs:icl}
\end{figure*}

Accessing databases via natural language queries (NLQs)  has been a long-standing goal of the database community~\cite{Li2014,Li2017,Quamar2022}.   Recently, LLMs with their superior capability of natural language understanding and code generation, have achieved great strides in the accuracy of Text-to-SQL generation as seen via public benchmarks~\cite{li2024can,spider2018yu}. The benchmark numbers are averaged over multiple database schemas, and typically evaluated zero-shot on unseen schema.  However, database-specific accuracy shows significant variation with changing schema.  In Figure~\ref{fig:zs:icl} we show accuracy on 11 schemas.  For some databases, the accuracy is low enough to be of little use in practical systems.  Such an experience is common for enterprises, whose databases are private to frontier LLMs.  Consequently, the enterprise may be willing to organize workload of NLQs with expert provided correct SQL, and adapt the LLM to increase its Text-to-SQL accuracy to acceptable levels.

Currently, there are two options for adaptation: fine-tuning and in-context learning.
Fine-tuning requires a lot of labeled data up-front, is unaffordable for small enterprises, and leads to forgetting of other tasks.  These limitations have caused great interest in non-fine-tuning based adaptation methods. In-context learning (ICL) is currently the go-to method for adapting an LLM to new databases without fine-tuning~\cite{Pourreza2024CHASESQLMR, wang2024macsql}. ICL just requires modifying the input prompt with a few examples of labeled related Text-SQL pairs to adapt the model on-the-fly for each user query.   Often, ICL leads to a significant boost in accuracy, as we show in Figure~\ref{fig:zs:icl}.

In this paper we propose methods that goes beyond in-context learning in harnessing labelled data of an enterprise.  Our work is based on two premises: \begin{enumerate*}
    \item The workload of an enterprise often contains clusters of highly similar SQLs.  
    We observed this pattern in the real workload of a very large bank, discussed more in Section \ref{sec:realW}.  The user's question (NLQ) could be different because of the inherent diversity of natural language, but the generated SQL is often highly similar to previously seen SQLs.  
    \item In-context learning disappoints in how well it harnesses the related queries.  We show two examples in Table~\ref{tab:icl-failure-examples}, where the LLM fails to generate the correct SQL even in the presence of an in-context example with SQL differing only by a constant.
\end{enumerate*}

\begin{table}[]
\centering
\begin{small}
\begin{tabular}{|p{0.95\columnwidth}|} \hline
\begin{minipage}{\linewidth}
Example 1: \\
Question: How many K-12 schools in Contra Costa register more than 420 free meals but have free or reduced-priced meals numbering under 610? 
\begin{lstlisting} 
SELECT COUNT(CDSCode) FROM frpm WHERE `County Name` = 'Contra Costa' AND `Free Meal Count (K-12)` > 420 AND `FRPM Count (K-12)`< 610
\end{lstlisting}
\end{minipage}
———————— \\
Question: In Los Angeles how many schools have more than 500 free meals but less than 700 free or reduced price meals for K-12? 
\begin{minipage}{\linewidth}
\begin{lstlisting}[mathescape, escapeinside=||]
SELECT COUNT(CDSCode) FROM frpm WHERE `County Name` = 'Los Angeles' AND `Free Meal Count (K-12)` > 500 AND `|\hlred{Enrollment (K-12)}|` < 700
\end{lstlisting}
\end{minipage} \\
\hline
\begin{minipage}{\linewidth}
Example 2: \\
Question: What is the total number of home team goals scored by Eric Djemba-Djemba? 
\begin{lstlisting}
SELECT SUM(t2.home_team_goal) FROM Player AS t1 INNER JOIN match AS t2 ON t1.player_api_id = t2.away_player_9 WHERE t1.player_name = 'Eric Djemba-Djemba'
\end{lstlisting}
\end{minipage}
———————— \\
Question: Aaron Lennon refers to player\_name = 'Aaron Lennon'; How many home team goal have been scored by Aaron Lennon? 
\begin{minipage}{\linewidth}
\begin{lstlisting}[mathescape, escapeinside=||]
SELECT SUM(t2.home_team_goal) FROM Player AS t1 INNER JOIN match AS t2 ON t1.player_api_id = t2.|\hlred{home\_player\_11}| WHERE t1.player_name = 'Aaron Lennon';
\end{lstlisting}
\end{minipage} 
\\
\hline
\end{tabular}
\end{small}
\caption{Examples where the ICL method generated wrong output in the presence of a very similar example. In the examples, the first NLQ-SQL pair is the ICL example, followed by the user NLQ and predicted SQL.}
\label{tab:icl-failure-examples}
\end{table}





We design a system \sysname, to go beyond in-context learning to harness 
closely related queries to improve accuracy of SQL generation. Our core idea is to convert previously seen labeled Text-SQL pairs to templatized Text-SQL forms to foster greater match to future queries.  When a future query matches one of the stored templates, \sysname\ generates the SQL via a dedicated template constrained decoding. 
We show that accuracy of matched queries jumps from approximately 60\% with ICL to almost 90\%  via our constrained decoding.
This implies that recurring queries can be executed with significantly higher reliability than baseline ICL methods. In addition, we take advantage of the templatized form to achieve almost a factor of two improvements in inference throughput.  

We encountered two primary challenges in the implementation of \sysname.  The first challenge was designing the matcher module for accurately matching queries to a stored pool of templatized Text-SQL pairs. We need to reject queries that do not match any template in the pool and choose the correct template from the pool for the rest. Since a wrongly matched template is guaranteed to provide the wrong SQL with constrained decoding, high accuracy in this step is crucial.  We propose three strategies to boost the accuracy of the matcher beyond simple thresholded cosine similarity of sentence embeddings:  (1) casting template match as a natural language inference problem, (2) masking parts of the user question, (3) generating multiple synthetic paraphrases of the original labeled Text-SQL pair to provide multiple text annotation to a shared template. Together, these strategies obtained a jump in selection/rejection accuracy from 73\% to 91\%
beyond the baseline.

The second challenge was efficiently and accurately enforcing the template constraints as the LLM generates the SQL for a user query.  Recently, many libraries have been developed for constraining text generated by LLMs to satisfy constraints specified as a regular expression or context-free grammar~\cite{geng-etal-2023-grammar, beurerkellner2024guidingllmsrightway, willard2023efficient}.  
We show how to cast the template as a flexible grammar that provides better agreement with the LLMs formatting biases by converting the masked SQL into a regular expression derived from the SQL grammar.   We present an efficient two-phase decoding algorithm for efficient constrained SQL generation.  In the first phase, we incur a one-time overhead to pre-compile the template grammar to generate LLM-aligned token sequence corresponding to the query-invariant part of the SQL template.  In the second phase, we efficiently fill in only the query-specific masked literals.  Overall, these lead to similar accuracy with decoding time reduced to $0.4-0.6\times$ of the library default.

\paragraph{Contributions}
\begin{enumerate*}
    \item Introducing the paradigm of template constrained decoding for recurring queries to address the accuracy and latency challenges of Text-to-SQL generation in an enterprise.
    \item Design of an accurate template matcher module that decides if natural language question corresponds to an SQL that conforms to one of the stored templates. 
    \item Accurate and efficient adaptation of grammar constrained decoding libraries for template constrained SQL generation.
    \item Evaluation on two SOTA Text-SQL benchmarks and five LLMs yielding an accuracy jump from an average of 60\% (with ICL) to almost 90\% (with \sysname) on a workload of recurring queries in the BIRD databases. 
\end{enumerate*}

\section{Related Work}
Text-to-SQL generation has been a very active and fast-progressing research area, with significant progress made in various areas, including design of models~\cite{ratsql2020wang,li2024codes}, schema and value subsetting~\cite{Li_Zhang_Li_Chen_2023,kothyari-etal-2023-crush4sql}, prompts and inference pipelines like CoT, consensus based reranking~\cite{Pourreza2024CHASESQLMR, Pourreza2023DINSQLDI, gao2023text,zhang-etal-2023-act,lee2024mcs}, and other reasoning-based methods~\cite{zhai2025excotoptimizingreasoningtexttosql}. 
%
%
These techniques contribute to improving the baseline performance of Text-SQL systems across schema.  For enterprises, where high accuracy is of paramount importance, and where memory of users and deployments are easily available, workload-drive customization of generic designs is of great interest.  We review existing methods of adapting pre-trained LLMs with schema-specific workloads.

\paragraph{Workload-driven Customization}
One class of methods proposes to fine-tune pre-trained LLMs to the Text-SQL tasks.  Examples include CodeS~\cite{li2024codes} that focused on fine-tuning for cross-schema generalization.  One challenge with fine-tuning models for a specific target database is collecting labeled data up front.  Some methods propose to augment with synthetic examples~\cite{victorialin2021,awasthi2022}.  However, with the advent of LLMs, the focus shifted to on-the-fly customization without model fine-tuning. LLMs naturally support In-Context Learning~\cite{gpt3brown2020language} where examples of Text-SQL pairs provided within the context have been found to adapt the LLM to SQLs of a target database~\cite{li2024codes,zhang-etal-2023-act,DBLP:journals/corr/abs-2502-14913}.  Other methods of adaptation include case base reasoning, and these have been found to be effective for generating SQL as a relational algebra tree~\cite{Varma2023}.

\textbf{Template-based generation} Templates have been harnessed for reducing the complexity of SQL generation in many prior systems.    ZeroNL2SQL~\cite{fan2024combining} uses a small language model to create candidate templates, and then uses a larger LLM to generate the final SQL via soft prompting and iterative feedback. 
CatSQL~\cite{fu2023catsql} uses a custom deep learning model to generate templates and fill the slots in the template, followed by semantic correction and post-processing of the generated SQL to fix the errors.  AmbiQT~\cite{bhaskar2023} harnesses templates for generating structurally diverse SQLs.   However, we are not aware of any prior work that proposes to harness existing labeled data as templates for more accurate SQL generation for similar queries.

\textbf{Constrained-decoding/grammar-guidance} Grammar- constrained decoding techniques ensure syntactically valid generation while leveraging the semantic understanding capabilities of large language models (LLMs). For SQL generation PICARD~\cite{picardScholak2021} pioneered this paradigm by integrating incremental parsing constraints during autoregressive decoding, dynamically rejecting tokens that violate SQL grammar rules and significantly improving execution validity. Recently, more general forms of constrained generation is supported by libraries such Transformer-CFG~\cite{geng-etal-2023-grammar}, DOMINO~\cite{beurerkellner2024guidingllmsrightway} and Outlines~\cite{willard2023efficient}. These differ in how they balance validity guarantees with computational efficiency through hybrid static/dynamic analysis of grammatical structures.  Transformer-CFG~\cite{geng-etal-2023-grammar}
introduced explicit modeling of context-free grammar (CFG) states through finite state machine (FSM) representations. This approach parses the entire vocabulary against the current FSM state, masks invalid token IDs, and updates the state based on sampled tokens - ensuring syntactic correctness at the cost of substantial inference overhead due to real-time FSM transitions and vocabulary-wide validity checks. 
%
Recent advancements like DOMINO~\cite{beurerkellner2024guidingllmsrightway} address this efficiency challenge through offline precomputation of prefix trees (tries) for each FSM state. By encoding valid token sequences in trie structures during preprocessing, DOMINO improves the latency of validity checks during decoding using tries, reducing computational complexity while maintaining grammatical constraints. Outlines~\cite{willard2023efficient} is another open-source Python library.
We are using Outlines as part of our system to generate SQL, but we innovate in how we efficiently invoke the library for template constraining.

\newcommand{\hy}{\hat{\vy}}
\section{Our approach}

\begin{figure*}[h]
    \centering
    \includegraphics[width=\textwidth]{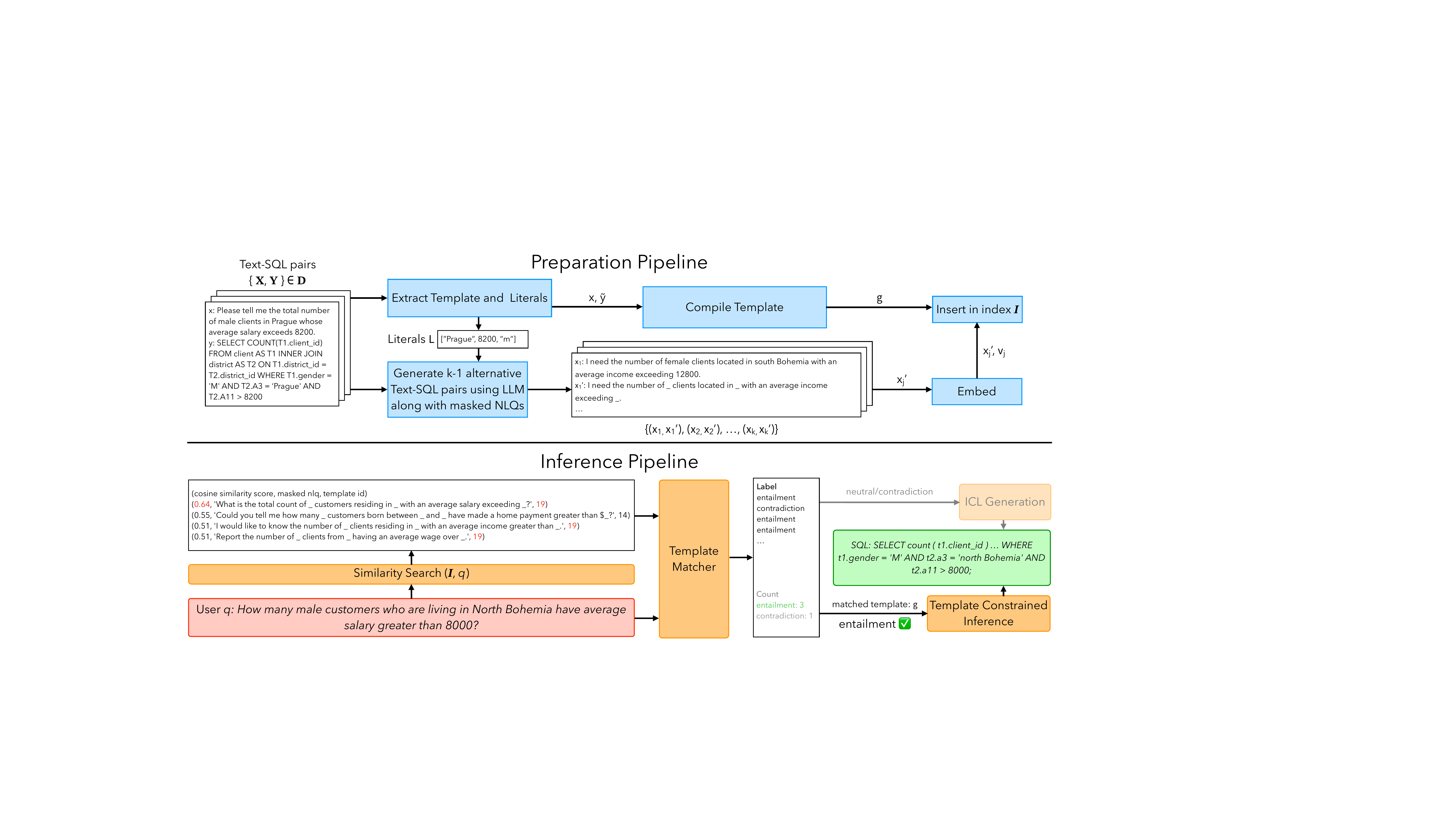}
    \caption{System Architecture of \sysname. The top part shows the processing done on labeled queries to extract and index templates. The bottom part shows steps during inference of each user query. See Figure~\ref{fig:tecod_inference} for Template Compilation and Template Constrained Inference.}
    \label{fig:system}
\end{figure*}

\paragraph{Problem Statement}
We are given a database DB with schema $S$ on which we wish to support natural language querying. Let $M$ denote an instruction-tuned LLM that given any user's natural language question (NLQ) $\textx$ and the schema $S$, can generate an SQL $\hy$ for $\textx$.   The default LLM may not provide high accuracy of conversion of NLQs to SQL.  We assume that there exists a workload of previous NLQs along with correct expert provided SQLs $D = \{(\vx^1,\vy^1),\ldots(\vx^N,\vy^N)\}$.  $N$ may not be large enough to perform supervised fine-tuning of $M$ to the schema and queries of this DB.  Our goal is to harness alternative strategies of customization.

\paragraph{A baseline method: In-Context Learning}
A baseline method is to retrieve labeled Text-SQL pairs from $D$ based on similarity of $\testx$ with corresponding text $\vx^i$ and include them as in-context examples in the prompt.  As we show in Figure \ref{fig:zs:icl}, including only a few related examples in the prompt can significantly enhance accuracy of the generated SQL.   However, while overall accuracy improves we also observe several cases where even when a highly related query is present in the context, the LLM fails to generate the correct SQL. Two examples appear in Table \ref{tab:icl-failure-examples}

Our proposed method \sysname\ is designed to more aggressively harness related examples in available labelled workload $D$.  

\paragraph{Overview of \sysname}
We present an overview of \sysname\ in Figure~\ref{fig:system}.  First, in the preparation phase we convert each (NLQ $\vx$,SQL $\vy$) pair in $D$ to a templatized form to allow better match.  These are inserted in an index \indexT\ for efficient retrieval in response to user queries. In Section~\ref{sec:prep} we provide details of this step.
%
%
During inference, given a user's natural language question (NLQ) $\testx$, \sysname\ first invokes a template matcher to decide if \indexT\ contains a template that conforms to the (unknown) correct SQL of $q$.   If a valid template is found, \sysname\ generates the SQL constraining it to follow the matched template. Otherwise, the SQL is generated using the standard method using in-context examples selected from $D$ or zero-shot depending on the baseline performance of the model.  The template selection and matching module is a critical component of this pipeline, and we describe its design in Section~\ref{sec:template}.  Another interesting component is how to decode by constraining as per the chosen template.  We describe the design of the template constrained decoding in Section~\ref{sec:decoding}.  


\subsection{Template Extraction and Indexing}
\label{sec:prep}

We process each NLQ $\vx$, SQL $\vy$ pair from $D$ into a templatized form in two steps.  First, we convert the SQL $\vy$ into a template $\ty$ and compile for efficient enforcement, and second we generate natural language annotations to the template $\ty$ so that these annotations can serve as search keys for matching with future natural language queries.  We describe each of these steps next:

\paragraph{Template Extraction and Compilation}
First, we convert the SQL $\vy$ into a template $\ty$ that is more likely to be shared by future queries. In this paper, we restrict the template to be the SQL with just constants and literals masked. Thus, a template in our definition is a parameterized SQL query. Two example SQLs and their corresponding templatized forms appear in Table~\ref{tab:example}.



Let L denote the set of literals present in SQL $\vy$. Extracting such literals from SQL is easy using an off-the-shelf parser like SQLGlot~\cite{sqlglot}.  
Next, we compile the templatized SQL into a flexible grammar $\vg$ to be used during constrained decoding.  More details of this step appear in Section~\ref{sec:decoding}.

\paragraph{Natural Language Annotations for Template}

\begin{small}
\begin{table*}[]
    \centering
      \caption{Example illustrating the templatization of Text-SQL Pairs after augmentation with synthetic NLQs}
    \label{tab:example}
    \begin{tabular}{|p{0.47\textwidth}|p{0.47\textwidth}|} \hline
    Original Text-SQL & Templatized and augmented \\ \hline
       NLQ $\vx$: How much, in total, did client number \hlred{617} pay for all of the transactions in \hlred{1998}? &  Masked NLQ 1: What was the total amount paid by client number \textcolor{red}{[number]} for all transactions in  \textcolor{red}{[string]}? \\
 & Masked NLQ 2: How much did client number  \textcolor{red}{[number]}  pay altogether for every transaction in \textcolor{red}{[string]}? \\
        & \\
         SQL $\vy$: SELECT SUM(T3.amount) FROM client AS T1 INNER JOIN disp AS T4 ON T1.client\_id = T4.client\_id INNER JOIN account AS T2 ON T4.account\_id = T2.account\_id INNER JOIN trans AS T3 ON T2.account\_id = T3.account\_id WHERE STRFTIME(\hlred{'\%Y'}, T3.date)= \hlred{'1998'} AND T1.client\_id = \hlred{617}
         & 
Template $\vy$: select sum(t3.amount) from client as t1 inner join disp as t4 on t1.client\_id = t4.client\_id inner join account as t2 on t4.account\_id = t2.account\_id inner join trans as t3 on t2.account\_id = t3.account\_id where strftime(\hlred{[string]}, t3.date) = \textcolor{red}{[string]} and t1.client\_id = \textcolor{red}{[number]}
\\ \hline
NLQ $\vx$: Who among \hlred{KAM}'s customers consumed the most? How much did it consume? &  Masked NLQ 1:  Which customer of \hlred{[string]} had the greatest level of consumption? What quantity did they consume? \\ 
& Masked NLQ 2: In terms of consumption, who stands out among \hlred{[string]}'s customers? How much did they consume? \\
& \\
SQL $\vy$: SELECT T2.CustomerID, SUM(T2.Consumption) FROM customers AS T1 INNER JOIN yearmonth AS T2 ON T1.CustomerID = T2.CustomerID WHERE T1.Segment=\hlred{'KAM'} GROUP BY T2.CustomerID ORDER BY SUM(T2.Consumption) DESC LIMIT \hlred{1} 
& 
Template $\vy$: select t2.customerid, sum(t2.consumption) from customers as t1 inner join yearmonth as t2 on t1.customerid = t2.customerid where t1.segment = \hlred{[string]} group by t2.customerid order by sum(t2.consumption) desc limit \hlred{[number]}
\\ \hline
    \end{tabular}
  
\end{table*}
\end{small}

To provide a covering set of annotations, we generate with the help of the LLM, synthetic Text-SQL pairs $(\vx_1,\vy_1), \ldots,(\vx_K,\vy_K)$
such that each SQL $\vy_i$ follows the template $\ty$.  This implies that each $\vy_i$ differs from the original SQL $\vy$ only in values of literals. 
Let $A(\ty)$ denotes the generated Text-SQL pairs including the original ($\vx,\vy)$.  We convert $\vx_i \in A(\ty)$ into a natural language annotation for $\ty$ by masking away tokens in $\vx_i$ that refer to constants literals  as follows.
Let $L$ be the literals in $\vy_i - \ty$ that we extract using an SQL parsing library. We mask the mentions of literals $L$ in the question to increase its match with future queries with differing literals.   Masking literal mentions in natural language is challenging. We use the following approach: We first sort the literals in $L$ by length in descending order so that longer literals are matched first. To mask out in $\vx_i$ the mention of a literal $\ell \in L$, we first look for case-insensitive exact matches of $\ell$ in $\vx_i$. After that, we do fuzzy matching using the RapidFuzz~\cite{max_bachmann_2024_10938887} Python package for approximate matching.  We denote the masked NLQ as $\tilde{\vx_i}$.   At the end of this process the masked $\tilde{\vx_i}$ should be relevant to $\ty$. Table~\ref{tab:example} shows examples of two Text-SQL pairs and their corresponding masked templatized forms.

Finally, we create vector embedding of all templated NLQ $\tx$s using a neural sentence embedding model like NV-Embed-v2~\cite{lee2024nv}.  The embedding is used to create the key to a hash-index with value as the template-id which in turn leads us to the corresponding compiled grammar.
We use \indexT\ to denote the index of masked NLQ and template-id pairs.


\subsection{Template Selection}
\label{sec:template}
Given a new user NLQ $\testx$, we need to find from \indexT\ a template $\ty$, if any, that would fit the correct SQL of $\testx$.  If no matching template is found, we default to the standard path of generating SQL using soft hints in the form of in-context examples selected from $D$.  If a matching template $\ty$ is found, we use the compiled grammar $\vg$ to generate the SQL using template constrained decoding as described in Section~\ref{sec:decoding}. Since this step forces the generation of the SQL to follow the prescribed template, it is important to perform the template matching and selection step with high accuracy.   The core module in this step is a template matcher model that we describe next.
\begin{algorithm}[]
\caption{Template Search with NLI Validation and Selection}
\label{alg:template-search-nli-selection}










\begin{algorithmic}[1]
\STATE \textbf{Input:} Natural Language Question ($q$), NLQ-Template Index \indexT, Top-k results ($k$)
\STATE \textbf{Output:} Template Matched  (\texttt{matched}), Template ID ($t\_id$)

\STATE \texttt{top\_k} $\leftarrow$ \texttt{sort(similarity\_search(}\indexT, $x$, $k$\texttt{), by=cosine similarity, desc)}
\STATE \texttt{best\_match} $\leftarrow$ \texttt{top\_k[0]}
\STATE \texttt{t\_id} $\leftarrow$ \texttt{\indexT[best\_match]}

\STATE \texttt{nli\_results} $\leftarrow$ [\texttt{NLI}($x$, \texttt{$\tx$}) \textbf{for} \texttt{$\tx$ in top\_k} \textbf{if} \texttt{\indexT[$\tx$]} == \texttt{t\_id}]

\STATE \texttt{nli\_label} $\leftarrow$ \texttt{majority\_vote(nli\_results)}
\STATE \texttt{matched} $\leftarrow$ (\texttt{nli\_label == 'entailment'})

\STATE \textbf{Return} \texttt{matched}, \texttt{t\_id}
\end{algorithmic}
\end{algorithm}


\paragraph{Template Matcher}  
The template matcher needs to decide if a user NLQ $\testx$ matches a stored template text $\tx$ such that the (unknown) SQL of $\testx$ would follow the template $\ty$. We found that pre-trained LLMs were not accurate for such reasoning. Also, just measuring the similarity of $\testx$ and $\tx$ using popular methods like cosine similarity of their respective sentence embeddings was not accurate enough. The task of deciding whether a template fits the correct SQL for an input NLQ requires more nuanced modeling.  Towards this end we trained a dedicated template matching model by repurposing 
a natural language inference (NLI) model.  An NLI model takes as input a pair of natural language sentences $s_1,s_2$ and decides if the logic in sentence $s_1$ entails, contradicts, or is neutral with the logic in sentence $s_2$.  Unlike embedding based models, state-of-the-art NLI models capture fine-grained interaction between the words of the two sentences using a Transformer with bidirectional attention to decide on these labels. In our application, one of the sentences $\tx$ is a NLQ with holes corresponding to the masked literals, and the other is the user NLQ $\testx$.  For example, in Figure~\ref{fig:system} we need to match the user NLQ $\testx$ correctly to the first, third and fourth masked annotation, and not the second. We fine-tuned a pre-trained NLI model to this form of the input.  Since we already generated alternative NLQs for a template, we create pairs out of these and call these positive pairs.  Next, for each NLQ, we found the closest NLQs from a different template, and call these as negative NLQs.  These examples across all templates are used to fine-tune existing pre-training NLI models like BERT\cite{devlin2019bertpretrainingdeepbidirectional}. We further increase robustness of the matcher to literal masking errors by training with both masked and un-masked NLQs.
We present further details of the training process, and the accuracy gains with a trained matcher model in the experiment section.

Further, to avoid invoking the template matcher model on every stored NLQ in \indexT, we first filter Top-K matched NLQs using cosine similarity.  The template-id with the closest masked NLQ based on cosine is filtered.  The NLI model is invoked on each masked NLQ of the template.  We get the NLI-label (entailment, neutral, contradiction) along with its probability for each pair. The final NLI label for a template is the majority of the predicted labels. The overall pseudocode occurs in Algorithm \ref{alg:template-search-nli-selection}. 

\begin{figure*}[h]
    \centering
    \includegraphics[width=\textwidth]{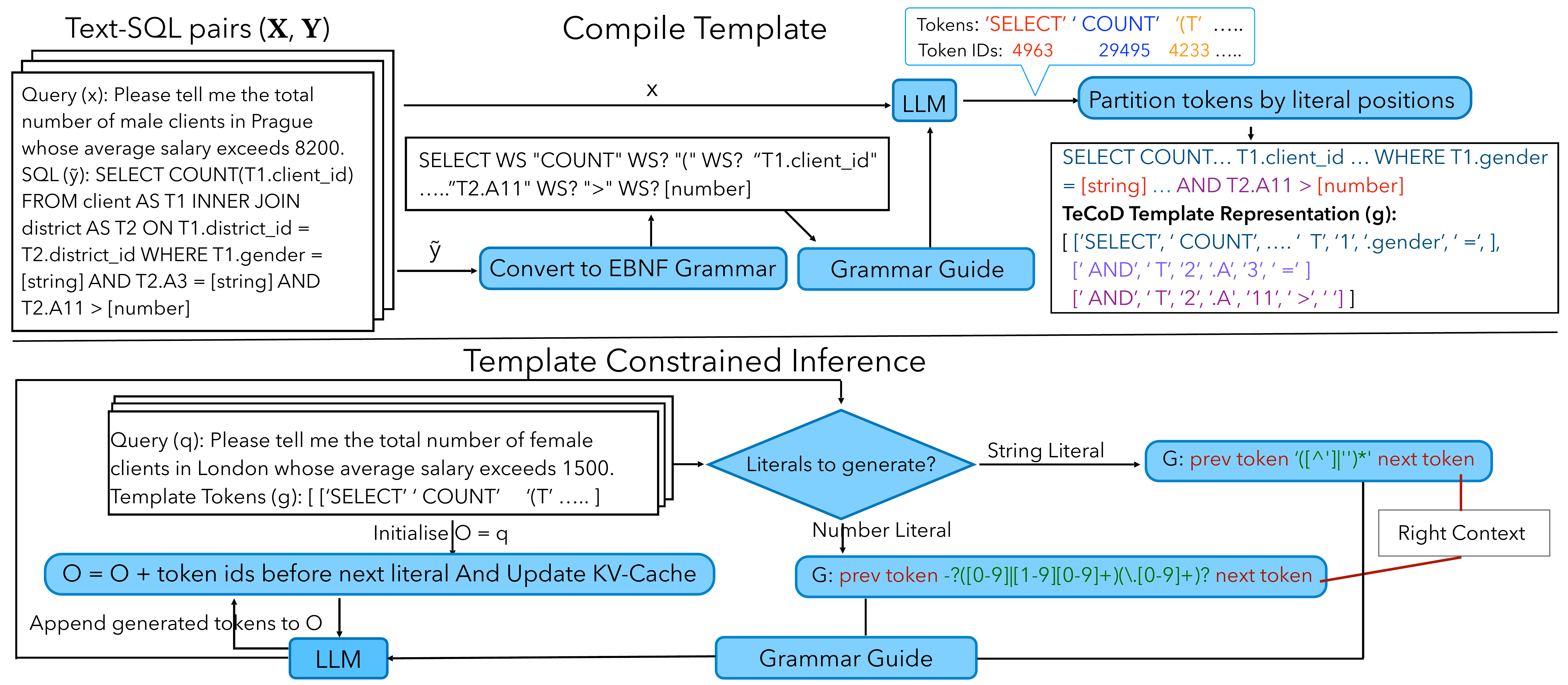}
    \caption{Template Compilation and Template Constrained Inference. Top part shows the one-time process of converting any template $\ty$ to a compiled representation $g$. Bottom part shows the iterative process to generate the template constrained SQL given user query $\testx$ and a selected template's grammar $g$.}
    \label{fig:tecod_inference}
\end{figure*}
\subsection{SQL Generation} \label{sec:decoding}
Based on the NLI label, we decide whether to do template constrained decoding or unconstrained generation.   In both cases, we include in-context-learning (ICL) examples in the form of demonstrations of pairs of user question and corresponding SQL.  
We next describe how we perform template constrained decoding.



We are given a selected template $\ty$ and a user NLQ $\testx$.  Our goal is to generate the SQL for $\testx$ while adhering to the template $\ty$.  Providing an LLM with template $\ty$ along with instructions to follow the template does not guarantee that LLM will always adhere to that template as we will show in the experiment section. Consequently, we modify LLM's decoding mechanism to be guided by the grammar-constrained decoding (GCD) rules. We present a brief background of how recent GCD methods work and then describe our adaptation to the template constrained generation task.

\paragraph{Background: Grammar constrained decoding (GCD) in LLMs}
Given an LLM $M$ and grammar $G$ which can be expressed as a regular expression or a context free grammar, recent libraries like Transformer\_CFG~\cite{geng-etal-2023-grammar} and Outlines~\cite{willard2023efficient} constrain LLM generated text to be valid as per the grammar $G$.  The LLM generates the text token-by-token using auto-regressive decoding.  At the $t$-th step of generation, let $y_1,\ldots,y_{t-1}$ denote the tokens generated so far, and let $P(y|y_1,\ldots,y_{t-1})$ denote the LLM's output probability distribution over its token vocabulary.  Normally, the LLM would sample a token with high probability from this distribution to get the next token, $y_t$.  Instead with GCD, the grammar $G$ is consulted to mask from $P(y_t|y_1,\ldots,y_{t-1})$ any token that is invalid as per the grammar $G$.   Generation stops when $y_t$ is a special end-of-sequence token. These libraries mask invalid token ids and maintain the state of allowed outputs efficiently while generating a token at each timestep.  

A template $\ty$ in our case is an SQL with zero or more empty slots to be filled with string or number literals. Here is a baseline method for harnessing the LLM to infer the slot values to generate the SQL for the user query $\testx$. Express the template as a regular expression such as shown in Table~\ref{tab:templates_snippets} (Row \#3), and simply invoke existing GCD libraries to generate the SQL.   A pseudocode is given in Algorithm~\ref{alg_fixed_g} as a reference. The statements in \textcolor{blue}{blue} show changes to the default LLM decoding algorithm to handle template constraints. 

\begin{algorithm}[]
\caption{Template Constrained SQL Generation}
\label{alg_fixed_g}
\begin{algorithmic}[1]
\STATE \textbf{Input:} LLM $M$, Grammar constraining library $C$, SQL Template $\ty$, Schema $S$, User query $\testx$
\STATE \textcolor{blue}{$\vg \gets {\text{Express\_As\_Grammar($\ty$)}}$}
\STATE \textcolor{blue}{$Guide \gets$ Initialize state of $C$ with $\vg$}
\STATE $O_1 = $ Start of sequence token.
\FOR{$t$=1 to Max-SQL-length}
\STATE $p(y) \gets$ Next token distribution from LLM $M(S,\testx,O_1\ldots,O_{t}$) \\
\STATE \textcolor{blue}{$m \gets$ Get mask of allowed tokens from $C(Guide,O_1\ldots,O_t)$} \\
\STATE $O_{t+1} \gets $ sample next token from $p(y)$ \textcolor{blue}{$\circ$ $m$} \\
\STATE If $O_{t+1}$ is EOS, Exit loop.
\ENDFOR
\RETURN Decode $O_1\ldots,O_t$ to SQL string.
\end{algorithmic}
\end{algorithm}

There are two problems with this approach: style mismatch and wasteful LLM invocations that we elaborate on next.

\paragraph{\textbf{Style Mismatch}} The specific formatting of the SQL used in the template may not be compatible with the SQL formats the LLM may have seen in its training corpus.  We found that each LLM has its own formatting and SQL styling preference such as case of keywords, use of aliases, punctuation, and white spaces between keywords.  In Table~\ref{tab:exampleAltSQL} we show multiple style in which two different LLMs generate the same SQL.
If we generate a fixed grammar that restricts the SQL generated to follow the string format in a given template, the LLM may not be accurate in generating the correct completions for the empty slots.  We address this limitation by converting the template into a more elaborate SQL grammar that captures all surface forms of the different but syntactically equivalent ways in which the same SQL can be represented.  An example of such a grammar is shown in Table ~\ref{tab:templates_snippets} under the name of Flexible Template.  Here we express the SQL template $\ty$ as a regular expression where SQL keywords, operators etc are replaced with corresponding non-terminals.  Every SQL keyword is allowed to be expressed in many different cases, and the whitespace between two keywords is flexible. Although tools like \texttt{sqlglot.qualify} can be used to normalize SQL queries by enforcing consistent aliasing, keyword casing, and formatting, we found that such normalization may not align with the LLM's SQL style. Such misalignment often leads to reduced accuracy when the LLM is used to fill the template.
We show that with the flexibility induced by the SQL grammar, the generated SQL accommodates the different surface forms (as shown in Table \ref{tab:exampleAltSQL}) in which the initial template is expressed. However, this flexibility comes with run-time overheads since for every generated SQL token the LLM needs to express its preference via its token distribution as shown in Algorithm~\ref{alg_fixed_g}.

\begin{table}[]
\centering
\begin{small}
\begin{tabular}{|p{0.98\columnwidth}|} \hline
\begin{minipage}{\linewidth}
\begin{lstlisting}[mathescape]
$\ty$: SELECT * FROM Office WHERE Name = [String] Limit [Number];
Grammar:
STRING_RULE: '([^'|'']*)'
SINGLEDIGIT: [0-9]
NUMBER_RULE: ("-"? (SINGLEDIGIT | [1-9] SINGLEDIGIT*)) ("." SINGLEDIGIT+)? ([eE] [+-]? SINGLEDIGIT+)?
WS: [ \t\n\r]
SELECT_RULE: "select" | "SELECT" | "Select"
FROM_RULE: "from" | "FROM" | "From"
WHERE_RULE: "where" | "WHERE" | "Where"
LIMIT_RULE: "limit" | "LIMIT" | "Limit" ....
\end{lstlisting}
\end{minipage}

\begin{minipage}{\linewidth}
\begin{lstlisting}[mathescape, escapeinside=||]
Fixed Template:
start: "SELECT" " " "*" " " "FROM" " " "Office" " " "WHERE" " " "Name" " " "=" " " [STRING_RULE] " " "Limit" " " [NUMBER_RULE] ";"
\end{lstlisting}
\end{minipage}

\begin{minipage}{\linewidth}
\begin{lstlisting}[mathescape]
Flexible Template:
start: WS? SELECT_RULE WS "*" WS FROM_RULE WS "Office" WS WHERE_RULE "Name" WS "=" WS STRING_RULE LIMIT_RULE NUMBER_RULE (WS? | ";"?)
\end{lstlisting}
\end{minipage}
\\
\hline
\end{tabular}
\end{small}
\caption{Two different grammars for a template $\ty$: (1) Fixed and (2) Flexible. Quoted strings (e.g.\texttt{"Name"}, \texttt{" "}) are fixed text to be emitted directly in the SQL output, while elements like \texttt{SELECT\_RULE} are grammar rules expanded during decoding.}
\label{tab:templates_snippets}
\end{table}

\begin{table}[]
\centering
\begin{small}
\begin{tabular}{|p{0.98\columnwidth}|} \hline
\begin{minipage}{\linewidth}
\begin{lstlisting}[mathescape, aboveskip=0pt, escapeinside=||]
Llama: SELECT song_name FROM singer WHERE |\textcolor{red}{AVG}| > ( SELECT AVG(age) FROM singer )|\textcolor{red}{;}|
\end{lstlisting}
\end{minipage}
\begin{minipage}{\linewidth}
\begin{lstlisting}[mathescape, escapeinside=||]
Granite: SELECT song_name|\textcolor{red}{\textbackslash n}|FROM singer|\textcolor{red}{\textbackslash n}|WHERE age > (SELECT |\textcolor{red}{AVG}|(age) FROM singer)|\textcolor{red}{;}|
\end{lstlisting}
\end{minipage}
\begin{minipage}{\linewidth}
\begin{lstlisting}[mathescape, escapeinside=||]
CodeS: SELECT song_name FROM singer WHERE age >(SELECT |\textcolor{red}{avg}|(age) FROM singer)
\end{lstlisting}
\end{minipage}
\\
\hline
\end{tabular}
\end{small}
\caption{This example shows differences in LLM formatting preferences such use of \texttt{\textbackslash n or space}, small or capital case for SQL function AVG, optional \textquotesingle;\textquotesingle , spacing around brackets.}
\label{tab:exampleAltSQL}
\end{table}

%

\paragraph{\textbf{Reducing LLM invocation cost via Partitioned Decoding}} In a template, typically a large portion stays the same across user queries, only the slots are potentially query dependent.  We address this limitation by designing a partitioned generation method that proceeds in two phases.
First, in a one-time template compilation phase,  we partition the template grammar $\ty$ into the parts that are independent of user query $\testx$ and masked literals that are query-specific.  We invoke the baseline whole grammar GCD method (Algorithm~\ref{alg_fixed_g}) once with the full grammar $G(\ty)$  and remember the token-id sequences for the static parts.   Second, at inference time, for each user query $\testx$  conditioned on the static token-ids, the LLM generates the literals using GCD with only the string or number literal as specified.  
One subtle challenge with such partitioned generation arises from how the LLM tokenizes strings. In a template like "select * from office where (name = [String]) Limit [Number];" it is clear that only the literals within the box bracket need query-specific constrained generation.  However, the LLM's preferred tokenization may straddle across the two partitions. For example the LLM may have created a single token "1;" but partitioned generation will not allow generation of tokens that straddle partition boundaries.  Such an issue appears in both the right and left context.    To handle this problem we apply a simple fix: we move boundary tokens from the static partitions as left and right context around the literals that are GCD generated.  We will show in Section~\ref{sec:temp_cons_decoding} that such contextualization during literal generation is crucial for accurate SQL generation.

The overall pseudocode appears in Algorithm~\ref{alg:efficient_decoding}. 
To avoid repeated full-template decoding, we first precompute token IDs for the template using Algorithm \ref{alg_fixed_g} which is a one time step and then partition them into static segments and literal slots $g$ (TeCoD Template Representation) as shown in figure \ref{fig:tecod_inference}. At inference, given a user query $\testx$, only the slot values are generated via GCD. For each literal slot, we initialize with left and right context tokens (extracted respectively in lines 11 and 12).  We then apply constrained decoding using regular expression for string or number literals as needed. The KV-cache from prior static tokens is reused for efficiency by avoiding repeat computation of key-value vectors. Table~\ref{tab:efficient-docoding-results} provides the latency comparisons between standard constrained decoding and our two-phase approach. We discuss  computational overhead of GCD in Section \ref{sec:temp_cons_decoding} where we compare latency overhead of two phase decoding with standard GCD and unconstrained generation.

Our approach builds on the strengths of CFG-based methods while optimizing for their computational inefficiencies for template-constrained generation. By narrowing the focus of the LLM to only the masked portions of the query and leveraging regex-guided decoding, we achieve a balance between efficiency, accuracy and practicality, which makes it particularly suitable for real-world applications with strict latency requirements. 



\begin{algorithm}
\caption{Efficient Partitioned Constrained Decoding}
\label{alg:efficient_decoding}
\begin{algorithmic}[1]
\STATE \textbf{Input:} LLM $M$, Grammar constraining library $C$, SQL Template $\ty$, Schema $S$, User query $\testx$
\STATE \textbf{Generating LLM Aligned Token ID Sequence (Offline):} 
\STATE $O_1\ldots,O_t \gets $ Invoke Algorithm \ref{alg_fixed_g}$(M, C, \ty, S, \testx)$ \\
\STATE $g \gets$ Partition By Literals($O_1\ldots,O_t$) \\
//\sysname\ Template Representation
\vspace{0.5em}
\STATE \textbf{Inference:}
\STATE $num\_regex \gets \texttt{-?([0-9]|[1-9][0-9]+)(\char92.[0-9]+)?} $ 
\STATE $str\_regex \gets \texttt{\textquotesingle([\char94\textquotesingle ]|\textquotesingle\textquotesingle)*\textquotesingle}$
\STATE $O \gets $ Tokens for $S + \testx$
\STATE $cache \gets \phi$  \hspace{5mm} //kv\_cache
\FOR{$i = 0$ to $\#literals-1$}
    \STATE $next\_token \gets $ Pop first token from $g_{i+1}$
    \STATE $prev\_token \gets $ Pop last token from $g_{i}$
    \IF{$literals_i$ is string}
        \STATE $guide_{str} \gets $ Initialize $C$ with $(prev\_token+str\_regex + next\_token)$ \\
        \STATE $O \gets $ LLM $M(O + g_i, guide_{str}, cache)$ \\
    \ELSIF{$literals_i$ is number}
        \STATE $guide_{num} \gets $
        Initialize $C$ with $(prev\_token+num\_regex + next\_token)$ \\
        \STATE $O \gets $ LLM $M(O + g_i, guide_{num}, cache)$ \\
    \ENDIF
    \STATE Update cache
\ENDFOR
\STATE $O \gets O + g_{\#literals}$
\RETURN $O$
\end{algorithmic}
\end{algorithm}


\section{Experiment Setting}
We evaluate the performance of \sysname\ in terms of both execution accuracy of the generated SQL and running time.  We consider two kinds of test queries: \matched\ where a matching template is present in $\tset$, and \unmatched\ where no matching query is present. Using these, our goal is to answer the following research questions using this empirical evaluation.
\paragraph{Research questions}
\begin{enumerate}[leftmargin=0.4cm, itemsep=0pt]
\item RQ0: Is there sufficient evidence of template reuse in real-life query workloads?
\item RQ1: For queries in the \matched\ set does \sysname\ provide any gains beyond existing option of adapting with in-context learning. 
\item RQ2: What is accuracy of template selection for \matched\ and \unmatched\ queries?
\item RQ3: What is the impact of various aspects of our template matching module --- masking literals, augmentation with synthetic NLQs, use of a fine-tuned NLI model.
\item RQ4: What is the accuracy boost with a flexible template enforcement beyond a string based template constraint? 
\item RQ5: What is the running time overhead of \sysname?
\end{enumerate}

\subsection{Dataset}
We present our evaluation on two popular Text-to-SQL benchmarks: BIRD-SQL~\cite{li2024can} and Spider~\cite{yu2019spider}.
In all cases we evaluate on the Dev set of the benchmark.
Unfortunately, these benchmarks are curated without any regard to preserve the actual frequency of occurrence of repeated queries in the workload. We handle this limitation by experimenting on two kinds of datasets as described below:

\subsubsection{Template with synthetic queries}
To simulate the workload in real enterprise databases where repeated query templates are commonplace (as we show in Section~\ref{sec:realW}), we augment each of these benchmarks as follows.  Let $T$ be a set of labeled Text-SQL pairs in a schema.  We first split $T$ into two equal halves randomly called matched $T_m$ and unmatched set, $T - T_m$. 

For each Text $\vx$ and SQL $\vy$ in the matched set $T_m$, we generate one synthetic SQL $\vy'$ and its corresponding NLQ $\vx'$ using OpenAI o3-mini.   The synthetic SQL $\vy'$ samples a different value of literal from the database compared to what is present in the original SQL $\vy$.  The corresponding NLQ $\vx'$ is a natural language utterance that the LLM provides of $\vy'$.  We further prompt the LLM to ensure that $\vx'$ is a sufficiently different paraphrase of $\vx$.  

Now the labeled pool $D$ available for indexing is the pairs ($\vx',\vy'$), whereas the original pairs $(\vx,\vy)$ are used to evaluate the LLM.  Thus, the whole of $T$ is used for evaluation, with the subset in the matched set guaranteed to find a matching template in $D$.

\subsubsection{Non-synthesized Template and Test Set}
\label{sec:overall_non_syn}
\label{sec:data_non_syn}
To provide a more comprehensive understanding of \sysname's practical performance and address concerns regarding potential biases introduced by this synthetic data augmentation, we conducted an evaluation on a test set and template pool created entirely out of real queries in the benchmark.  We found only a small number of recurring SQL templates in this workload.
In the BIRD-dev set, we found 61 templates covering 134 queries out of 1534, and in the Spider-dev set 482 templates covering 966 questions out of 1034. We evaluate our system with this subset of 61 and 482 queries as our template pool, and use it to create our template bank along with paraphrases generated on these questions. The rest of the questions that are covered by the template make up our test set. Our test set on Bird is comprised of 73 matching queries across databases, and we sampled an equal number of non-matching queries to make the split even, making it a total of 134 queries. On Spider, we have a test set of 499 queries, of which 484 are matching and 15 are non-matching queries. The split is not even in the case of Spider, as most of the queries are covered by the template, and a few databases have no templates, and similarly for Bird, one database has no templateable questions. Please refer Table \ref{tab:dataset_stats}  in Appendix 
 \ref{app:nl_similarity} for more details.

\paragraph{\textbf{Evaluation Metrics}}
We use a widely adopted metric, Execution Match Accuracy (ExM). The ExM metric evaluates whether the predicted SQL and the gold SQL yield the same execution results on the database. 
Further, we use \textbf{Inference Latency}, to measure the efficiency of the decoding pipeline and the throughput improvements over the baseline.

\subsection{\textbf{Large Language Models}}
\label{sec:models}
We evaluate our system on general purpose models like Llama-3.1-8B-Instruct~\cite{grattafiori2024llama} and Granite-3.1-8B-Instruct~\cite{ibm_granite_2024_instruct}, and supervised fine-tuned models from the CodeS~\cite{li2024codes} series with 1B and 15B parameter variants, CodeS-1B-Bird-with-evidence, CodeS-1B-Spider, CodeS-15B-Bird-with-evidence, and CodeS-15B-Spider. 
We extend our evaluation
to include recent SOTA models for the Text2SQL task such as XiyanSQL Qwen-
coder (7B and 14B) \cite{XiYanSQL} and Arctic-R1-7B \cite{yao2025arctictext2sqlr1simplerewardsstrong}, which are publicly available, provide SOTA results on BIRD Single-Model Leaderboard, and comparable in scale to the models used in our current experiments. Of these Arctic-R1-7B is a reasoning model and is almost an order of magnitude slower than the rest of the models.  We will see that our method is orthogonal to the baseline model used, and provides gains across all LLMs. We did not include other recent models like DIN-SQL ~\cite{Pourreza2023DINSQLDI}  or CHASE-SQL~\cite{Pourreza2024CHASESQLMR}  because either the code is not publicly available or we included others like QwenCoder which are established better.

\paragraph{SQL Generation using LLM}
We prompt the LLM for generating SQL using an LLM specific prompt. In each case, there is a generic natural language instruction, followed by a description of the schema and metadata of the database queried, followed by the in-context examples, and then the current test question $\testx$.  However, instead of providing the entire database metadata, we filter the meta data as described below.
\paragraph{Schema Filtering}
Enterprise databases are comprised of a large number of tables and columns per table. If we include complete schema in the prompt, it will exceed the model’s context length. We use the models developed by CodeS~\cite{li2024codes} to do schema subsetting. 
Essentially, at inference time, we input database schema and NLQ to get relevance scores and using those scores, we choose top-k tables and columns, which would become part of the schema prompt.
\paragraph{Value Retriever}
In order to inform the LLM of the values of categorical columns that might possible match user NLQ, it is necessary to also retrieve candidate matching values from the database.
Suppose the question is “Who is the last F1 winner of the Sepang GP?” there has to be a correct linking between “Sepang” and the city column of the table that contains the winner's data. Here, again, we reuse the models in the CodeS repository to retrieve values. They proposed a coarse-to-fine-grained matching approach, where first, they use the BM25 index for the initial search and later use the longest common subsequence (LCS) algorithm to find the most relevant values. 

\paragraph{Grammar Constrained Decoding} For constraining the output of the LLMs, we are using Outlines~\cite{willard2023efficient} Python library.


\subsection{Training the Template Matcher} \label{sec:nli-training}
We fine-tuned the distilbert-base-uncased~\cite{Sanh2019DistilBERTAD} model for the NLI task using the training split of the BIRD benchmark (See ~\ref{app:nli_model} for the details on the embedding and NLI models used with non-synthesized workload). We generated 20 alternate NLQs per question from the train set using OpenAI o3-mini to serve as positive pairs.  To train the model, we needed hard negatives, which we mined using cosine similarity and added an equal number of negative pairs (with unmasked NLQ). Also, we added the same number of positive and negative pairs with masked NLQ to make the model more robust.
Note, that the template matcher model database-agnostic and is shared across databases unlike the template index.  The databases from which queries are sampled  during training are disjoint from the databases used during testing.   The model is able to generalize because the task is much simpler --- establishing the semantic  correspondence between user NLQ and masked-NLQs (from \indexT). The BIRD train provides a large number of examples.  Training on BIRD’s train set does not give an unfair advantage on BIRD dev set because the databases in these splits are entirely different.

For each database schema we create a different template index since our template refer to table and column names of a schema. Our system is designed for harnessing the previously seen queries of a specific enterprise database, to improve SQL generation performance of frequent queries.

\section{Results}
\begin{figure}[h]
    \centering
    \includegraphics[width=0.5\textwidth]{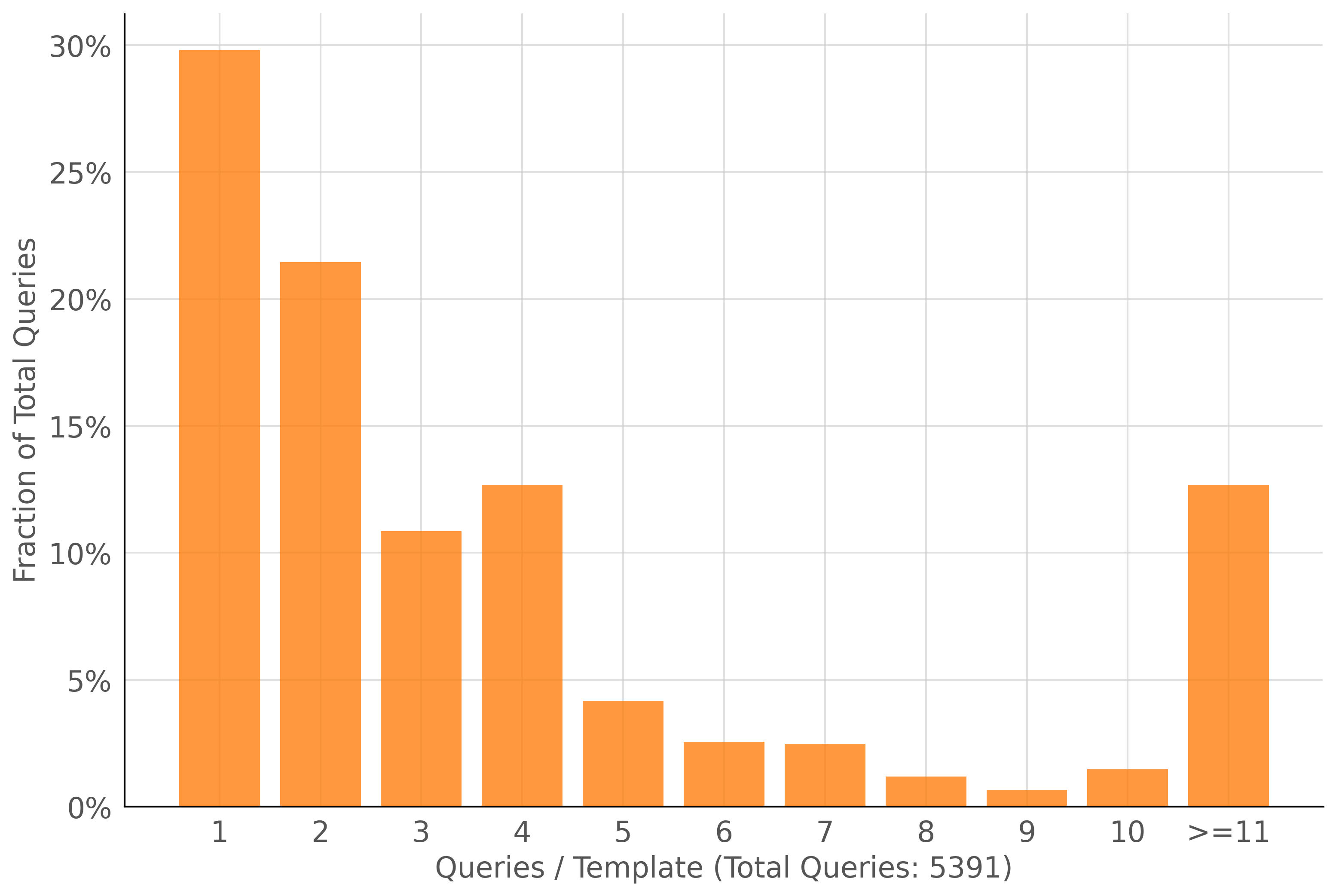}
    \caption{Workload distribution of a large bank. The $X$-axis shows the size of each template in terms of number of queries in that template.  The $Y$-axis shows the fraction of total queries that are present in templates of that size. This chart shows that 30\% of the queries are in templates of size 1, and 22\% in templates of size 2.  We show that more than 50\% of the queries would find a matching template in prior queries if queries arrive sequentially in the above workload.}
    \label{fig:real-workload}
\end{figure}

\subsection{Role of Templates in Real workloads}
\label{sec:realW}
As discussed earlier, standard benchmarks like BIRD and SPIDER do not preserve the occurrence frequency of repeated queries that is typical in large enterprises.  We were able to access the proprietary workload of a large bank in a large country. The workload reflects natural language queries submitted by business executives to the DBAs of the main data warehouse of the bank. Bank queries are OLAP in nature. Most of the queries are for analysis, reporting, or business intelligence purposes. We grouped queries based on whether they follow the same parametric template. Each group defines a template, and we define the size of a template as the number of queries that fall in that group.  In Figure~\ref{fig:real-workload} we show for each template size, the fraction of the total query workload that are part of a template of that size.  Observe that 30\% of the queries in the workload belong to a template of size 1, and about 22\% to templates of size 2 etc.  From these numbers, we can estimate that if queries in the workload were to arrive sequentially then for more than 50\% of the queries, a previous query with a matching template would have been observed.  In this case, every generated SQL is manually inspected by the DBA, so all previously occurring queries would be paired with a verified correct SQL.  With our system for more than 50\% of the queries, the matched template could be used to automatically generate a correct SQL with significantly higher accuracy than a baseline Text-to-SQL system.

\sysname\ is built on the premise that while natural language phrasing is diverse, the underlying SQL structure for common business queries is often recurrent. While the effectiveness of our system relies on having coverage for a given query in the template index, it is specifically designed to target these high-frequency head queries. As demonstrated by the enterprise workload analysis, where over 50\% of queries could find a matching template upon sequential arrival, TeCOD provides a pragmatic path towards high-accuracy SQL generation for the most common analytical needs, thereby significantly enhancing reliability for the bulk of an organization's day-to-day data interaction.

\subsection{Overall Text-to-SQL Execution Accuracy} \label{sec:overall}

\begin{table}[h]
\caption{BIRD and Spider ExM\% accuracy for various methods on various LLMs (CodeS-Bird-with-evidence for the BIRD and CodeS-Spider for the Spider dataset). Our method (TeCoD-GCD) that performs template constrained decoding provides significantly higher accuracy than the next best method ICL-3, for both synthesized and non-synthesized matched sets.}
\label{tab:merged_results}
\begin{tabular}{ll|r|r|r|r}
\toprule
\multirow[b]{2}{*}{Model} & \multirow[b]{2}{*}{Method} & \multicolumn{2}{c|}{Syn} & \multicolumn{2}{c}{Non-Syn} \\
\cmidrule(lr){3-4} \cmidrule(lr){5-6}
& & BIRD & Spider & BIRD & Spider \\
\midrule
\multirow[t]{4}{*}{CodeS-1B} & ZeroShot & 40.91 & 70.33 & 63.01 & 72.11 \\
& ICL-3 & 50.65 & 79.37 & 68.49 & 78.10 \\
& TeCoD-SGC & 54.55 & 82.32 & 64.38 & 79.75 \\
& TeCoD-GCD & 86.88 & 98.04 & 83.56 & 95.04 \\
\cline{1-6}
\multirow[t]{4}{*}{Granite-3.1} & ZeroShot & 26.23 & 64.44 & 43.84 & 66.74 \\
2B-Instruct & ICL-3 & 62.99 & 86.44 & 71.23 & 82.23 \\
& TeCoD-SGC & 74.94 & 94.70 & 79.45 & 89.46 \\
& TeCoD-GCD & 86.88 & 98.43 & 84.93 & 95.87 \\
\cline{1-6}
\multirow[t]{4}{*}{Llama-3.1} & ZeroShot & 44.81 & 77.01 & 61.64 & 75.00 \\
8B-Instruct & ICL-3 & 73.90 & 93.52 & 83.56 & 89.46 \\
& TeCoD-SGC & 84.16 & 94.30 & 87.67 & 91.32 \\
& TeCoD-GCD & 89.22 & 96.27 & 89.04 & 94.01 \\
\cline{1-6}
\multirow[t]{4}{*}{Granite-3.1 } & ZeroShot & 40.26 & 71.12 & 61.64 & 73.14 \\
8B-Instruct & ICL-3 & 62.47 & 87.03 & 84.93 & 84.09 \\
& TeCoD-SGC & 78.31 & 89.98 & 87.67 & 86.98 \\
& TeCoD-GCD & 89.22 & 98.62 & 89.04 & 95.45 \\
\cline{1-6}
\multirow[t]{4}{*}{CodeS-15B} & ZeroShot & 51.30 & 81.34 & 68.49 & 81.40 \\
& ICL-3 & 64.29 & 88.02 & 78.08 & 87.40 \\
& TeCoD-SGC & 69.35 & 88.61 & 78.08 & 86.98 \\
& TeCoD-GCD & 88.70 & 97.25 & 90.41 & 94.63 \\
\midrule
\multicolumn{6}{c}{\textbf{SOTA LLMs for Text-to-SQL}} \\
\midrule
\multirow[t]{4}{*}{XiYanSQL} & ZeroShot & 54.10 & 90.37 & 75.34 & 90.70 \\
QwenCoder & ICL-3 & 77.79 & 95.48 & 84.93 & 94.63 \\
7B-2504 & TeCoD-SGC & 81.95 & 94.50 & 89.04 & 92.98 \\
& TeCoD-GCD & 89.48 & 97.45 & 87.67 & 95.66 \\
\cline{1-6}
\multirow[t]{4}{*}{XiYanSQL} & ZeroShot & 55.66 & 94.30 & 73.97 & 93.80 \\
QwenCoder & ICL-3 & 79.87 & 97.45 & 89.04 & 96.49 \\
14B-2504& TeCoD-SGC & 90.39 & 98.43 & 93.15 & 97.31 \\
& TeCoD-GCD & 88.05 & 99.02 & 90.41 & 96.07 \\
\cline{1-6}
\multirow[t]{4}{*}{Snowflake} & ZeroShot & 63.25 & 89.00 & 79.45 & 88.02 \\
Arctic & ICL-3 & 69.22 & 93.52 & 79.45 & 92.15 \\
Text2SQL & TeCoD-SGC & 75.32 & 94.70 & 83.56 & 92.77 \\
R1-7B & TeCoD-GCD & 86.36 & 97.05 & 89.04 & 93.60 \\
\bottomrule
\end{tabular}
\end{table}

In Table~\ref{tab:merged_results} we compare the execution accuracy of the following methods on eight LLMs of different sizes and recency:
\begin{enumerate}[leftmargin=0.3cm, itemsep=0pt]
    \item ZeroShot: Where no in-context examples are provided in the prompt.
    \item ICL-3: Where three labeled Text-SQL pairs from $D$ whose NLQ is most similar to $\testx$. 
    \item \sysname-SGC: Here we match templates using \sysname, but instead of hard enforcement of the matched templates using GCD, we prompt with an additional instruction to the LLM to generate the SQL for $\testx$ in accordance with the template $\ty$.
    \item \sysname: Here we follow the full pipeline as outlined in Figure~\ref{fig:system}. 
\end{enumerate} 
We did not consider fine-tuning with the queries used for template construction because the number of queries is small.  Importantly, our setting assumes templates arrive over time, not all at once. Fine-tuning would require repeated retraining, which is impractical. Also with limited data, fine-tuning risks overfitting, forgetting, and other side effects. Moreover, FT is not always feasible in multi-tenancy models. 

\noindent
In Table~\ref{tab:merged_results} we include results of all four datasets as described. However, comparisons across datasets is not meaningful since the Synthetic and Non-Synthetic versions have very different test-sets as seen in Table~\ref{tab:dataset_stats}. Here the workload comprises of only queries for which a matching template exists in $D$, however, the identity of the matching template needs to discovered. We present results over a mix of matched and unmatched queries in Section~\ref{sec:unmatched}.

\noindent
In Table~\ref{tab:merged_results} comparing among the four methods across dataset-LLM combination, we can make these interesting observations.
\begin{enumerate}[leftmargin=0.3cm, itemsep=0pt]
\item Zero-shot accuracy is quite poor on the BIRD benchmark across all LLMs.  The Spider benchmark is considered easier, and there ZeroShot accuracy is much higher.
\item In-context examples provide improvements for all dataset-LLM combinations.  
Note, the selected ICL examples likely include one example with the matching template, and even then the accuracy boost is modest for most LLMs.
\item Once \sysname's template selection module chooses a template, and we instruct the LLM to follow the template (\sysname-SGC method), we observe a huge jump in accuracy beyond ICL-3.
\item However, \sysname-GCD  with strict enforcement of the chosen template provides significantly greatest overall gains, particularly for smaller models like CodeS-1B and Granite-2B.  In most cases, we achieve more than 89\% accuracy, and there is a huge jump in accuracy over \sysname-SGC.
\item As seen in the bottom part of the table, recent LLMs specifically trained for SQL generation provide much higher zero-Shot and few-shot accuracy than earlier models, but even on these TeCoD is able to provide gains.
While TeCoD-GCD generally demonstrates superior accuracy for matched queries compared to TeCoD-SGC across various LLMs and datasets, a nuanced observation arises when evaluating newer, state-of-the-art models on this  data. 
For highly capable LLMs like XiYanSQL QwenCoder-14B, which possess robust inherent understanding and SQL generation capabilities, the "soft guidance" provided by TeCoD-SGC may suffice. These advanced models might be proficient enough to adhere to the template without strict grammar enforcement, potentially remedying the errors of template selection. 
However, the latency of these newer models could be prohibitive, with latencies of 6.94s for XiYanSQL QwenCoder-7B and 9.67s for XiYanSQL QwenCoder-14B. In contrast, TeCoD-GCD delivers close enough accuracy on Granite-8B, which is significantly more responsive with a latency of just 3.64s.
\end{enumerate}

In summary, TeCoD not only elevates
weaker and smaller base models (like Granite-3.1-8B-Inst) to match performance of larger, stronger baselines (e.g., QwenCoder-14B ICL-3), but it also improves the performance of larger models. 
This positions TeCoD as a
generalizable and impactful inference enhancement method
for Text-to-SQL tasks.


\paragraph{Error Analysis}
While \sysname\ achieves strong execution accuracy across datasets, there remain a small number of cases (8-12\%) where the generated SQL is incorrect.  We present a brief analysis of the reasons for these errors.  
First, we account for the errors due to not being able to identify the correct template. Second, the bulk of the error is due to the wrong literal generation even after matching with the correct template.
We analyze the nature of these errors. 

\begin{enumerate}[leftmargin=0.3cm, itemsep=0pt]
\item A common source of error is when the literal name is an obscure entry (e.g., domain-specific jargon or an internal abbreviation) in the database, and the schema subset shown to the LLM prompt fails to retrieve the literal name.   
\item Another source of errors is number literals arising out of LLMs difficulty with numerical reasoning. An example is shown below: 
    \begin{lstlisting}[mathescape, escapeinside=||]
Gold: SELECT T1.frequency, T2.k_symbol ... WHERE T1.account_id = 3 AND T2.total_amount=3539
Pred: SELECT T1.frequency, T2.k_symbol ... WHERE T1.account_id = 3|\textcolor{red}{539}| AND T2.total_amount=3539
    \end{lstlisting}
\item A third source of errors is failure of grammar constrained decoding to terminate string literals particularly when the string itself contains a quote.
    Such behavior is especially noticeable in queries containing non-ASCII characters in string literals.
    \begin{lstlisting}[mathescape, escapeinside=||]
str_regex: $\texttt{\textquotesingle([\char94\textquotesingle ]|\textquotesingle\textquotesingle)*\textquotesingle}$
Gold: SELECT type FROM sets WHERE code IN ( SELECT setCode FROM set_translations WHERE translation='|{Huiti\`eme 'edition}|' )
Pred: SELECT type FROM sets WHERE code IN ( SELECT setCode FROM set_translations WHERE translation='|{Huiti\`eme \'edition}|'|\textcolor{red}{') GROUP BY TYPE; SELECT TYPE FROM sets WHERE code = 'n' )}|
    \end{lstlisting}
    Here, after generating tokens for the string literal, the LLM outputs \texttt{\textquotesingle\textquotesingle} instead of \texttt{\textquotesingle} or \texttt{\textquotesingle)}, and the grammar considers it as escaped single quote. As a result, the LLM can continue generating any token permitted by the string regex which includes all possible characters until stopping criteria is reached.
\end{enumerate}
We present more examples of errors in Section~\ref{app:error} of the Appendix.

\subsection{Impact of \sysname\ on queries without matching templates}
\label{sec:unmatched}
In TeCoD, when the template matcher incorrectly identifies a template for an unmatched NLQ, accuracy of those queries drop. 
We observed that zero-shot accuracy is the same roughly across matched (M) and unmatched (U).  ICL causes accuracy to jump by almost 20\% on an average for the matched set (M), while providing only a modest 2\% gains for the unmatched set.  With TeCoD, accuracy drops by between 3--7\% for the unmatched set, but because of the dramatic jump in accuracy of the matched set, the overall average accuracy improves. We observe jumps by ~15-36\% for BIRD-dev over ICL-3 and ~4-18\% jump for Spider-dev across LLMs.  In addition to the improved accuracy we also observe 2.2x lower latency. TeCoD-GCD is particularly useful in reducing the latency of reasoning models like Arctic Text2SQL where the baseline model takes 20 seconds per query on average, which reduces to 6 seconds.  Even though with TeCoD, accuracy drops by a small amount for the unmatched set, but since the baseline accuracy for this set is already low, (average accuracy of 50\% for BIRD), the practical impact of such a drop may be limited. Also, as an enterprise collects more queries to the template pool, the unmatched pool size is expected to shrink.

\subsection{Template Selection Accuracy}
\label{sec:temp_sel_accuracy}

\begin{table}[]
\setlength\tabcolsep{2.0pt}
    \centering
        \caption{Accuracy of template selection and rejection on BIRD and Spider datasets for baseline, our method, and various ablations on our method.  Best accuracy provided by our NLI-based template matching module without masking the user question.  Baseline method based on cosine similarity of question and template embeddings provides much worse matches.}
    \label{tab:selection}
    \begin{tabular}{|l|r|r|r||r|r|r|} \hline
     Dataset & \multicolumn{3}{c||}{BIRD} & \multicolumn{3}{c|}{Spider} \\ \hline
     Method    &  Selection & Rejection & Average & Selection & Rejection & Average \\ \hline
     Baseline    &  63.51 & 82.85 & 73.14 & 74.66 & 75.24 & 74.95 \\
     Ours, No mask & 92.21  & 89.01 & \textbf{90.61} & 97.64 & 86.86 & 92.17 \\
     Ours, With Masking  &  91.82 & 88.87 & 90.35 & 97.84 & 87.05 & \textbf{92.36} \\
     Ours Gold Masking & 91.69 & 88.61 & 90.16 & 97.64 & 86.86 & 92.17 \\   
     Ours, 1 NLQ & 87.14  & 87.43 & 87.29 & 92.93 & 84.38 & 88.59 \\
     Ours, No FT & 68.31  & 83.77 & 76.01 & 87.23 & 82.29 & 84.72 \\
     Ours, No FT, No mask & 30.91 & 95.29 & 62.97 & 56.39 & 89.33 & 73.11 \\
     Ours, No NLI & 71.43  & 89.66 & 80.51 & 92.34 & 85.71 & 88.97 \\ 
     \hline
    \end{tabular}
\end{table}

We next present the efficacy of our template selection module described in Section~\ref{sec:template}.  To bring out the merit of different design decisions in that module, we present a comparison with a number of ablations and baseline. In each case, we measure two kinds of accuracy: (1) Selection accuracy for queries where a matching template is known to be present in \indexT.  Errors in this case could be either because of deciding that no matching template is present, or choosing the wrong template.    (2) Rejection accuracy for queries where no matching template is present.  For such queries, the correct output is to reject all templates in \indexT.  We have an equal number of matched and unmatched queries, and also present Overall accuracy as the average of the two.

Table~\ref{tab:selection} compares the following methods:
\begin{enumerate}[leftmargin=0.3cm, itemsep=0pt]
    \item Baseline: A baseline method for template matching is to compare the embedding of the template with that of the NLQ using established methods like Cosine similarity.  Let $\ty$ be a candidate template, and we need to decide if a user question $\testx$ would lead to an SQL with template $\ty$.  In this method, this decision is made based on whether Cosine similarity of the embedding of $\tx,\testx$ is greater than a threshold $\eta$. To calculate the threshold, we create a sample set of questions with an equal number of positives and negatives by choosing a positive and a negative example per question from the synthetically generated Text-SQL pairs.  
The threshold is chosen to maximize the difference between the true positive rate and the false positive rate, giving the best balance between true positives and false positives.
\item \sysname's template matcher as described in Section~\ref{sec:template} where we use a fine-tuned NLI model. Details about NLI model training appear in Section~\ref{sec:nli-training}. We use the same model for Spider and BIRD.   Observe that our matching module provides significantly higher accuracy for both selection and rejection of templates.  compared to baseline of 73.14\%, we achieve an accuracy of 90.61\%.
\item Ours, with Masking:
In this version, we also mask literals in each arriving user query $\testx$ by first generating an SQL using the default LLM, and then fuzzy masking the literals in the SQL from $\testx$.  This method of masking the literal in the user question entails an additional overhead of generating the SQL using a default method.  In any case, there is no guarantee that the generated SQL is correct.  We observe that the accuracy stays more or less the same compared to our default no-masking approach.  One reason for this robustness is that we trained the NLI model with a mix of masked and unmasked user questions.
\item Ours with Gold Masking: In order to firmly establish the role of masking, we consider an oracle setting, where we use the gold SQL of the user NLQ $\testx$ to extract the literals, and mask them in the user question.  Masking with gold literals improves accuracy, but only slightly.  Based on these experiments we resolved to not mask literals in \sysname's final pipeline.
\item Ours, Single NLQ per template: We next study the impact of including multiple synthetic NLQs with template.  With just a single NLQ per template, the accuracy drops by almost 3.32\% drop compared to with 10 NLQs.   
    \item Ours with untuned NLI model: 
    When using the untuned NLI model (we used HF tasksource/deberta-base-long-nli) accuracy dropped significantly from 90.35\% to 76.01\%.
    \item Ours with untuned NLI model for unmasked NLQ: When the untuned NLI model is applied on umasked user questions, the drop in selection accuracy is drastic going from 92\% to 31\%.  This shows that our strategy of fine-tuning the NLI model with a mix of masked and unmasked user question was essential to enable the NLI model to perform well even on unmasked user questions.
\end{enumerate}

\begin{table*}[htbp] 
\centering 
\begin{small}
\caption{Execution Match(EX) and inference latency for different methods of template constrained generation on BIRD and Spider dev sets. For reference we also show the latency of unconstrained generation. Our method \sysname\ provides almost the same accuracy as using the full \GCD\ constraints, while reducing running time by about half compared to library default. CodeS-1B/15B is used for both BIRD and Spider, but refers to separate LLMs finetuned on each dataset's training set.}
\label{tab:efficient-docoding-results}
\setlength{\tabcolsep}{3pt} 
\begin{tabular}{llrrrrrrr}
\toprule
& \multicolumn{8}{c}{BIRD} \\
\cmidrule(lr){2-9}
\multirow{1}{*}{Method} & \multicolumn{2}{c}{CodeS-1B} & \multicolumn{2}{c}{CodeS-15B} & \multicolumn{2}{c}{\begin{tabular}{@{}c@{}}Llama-3.1 8B \\ Instruct\end{tabular}} & \multicolumn{2}{c}{\begin{tabular}{@{}c@{}}Granite-3.1 2B \\ Instruct\end{tabular}} \\
\cmidrule(lr){2-3} \cmidrule(lr){4-5} \cmidrule(lr){6-7} \cmidrule(lr){8-9}
& EX (\%) & Latency & EX (\%) & Latency & EX (\%) & Latency & EX (\%) & Latency \\
\midrule
Unconstrained & 38.07 & 1.00$\times$ & 48.57 & 1.00$\times$ & 39.18 & 1.00$\times$ & 24.12 & 1.00$\times$\\
\sysname & 91.40 & 1.15$\times$ & 92.44 & 0.87$\times$ & 92.18 & 1.24$\times$ & 88.07 & 0.62$\times$\\
\sysname\ (No Two-phase decoding) & 91.85 & 2.66$\times$ & 92.89 & 1.34$\times$ & 92.50 & 2.74$\times$ & 88.33 & 1.44$\times$\\
\sysname\ (No context) & 84.22 & 1.83$\times$ & 85.59 & 1.21$\times$ & 79.47 & 1.59$\times$ & 73.53 & 1.60$\times$\\
\SCD\ (No context) & 69.17 & 2.14$\times$ & 71.97 & 1.45$\times$ & 35.66 & 1.83$\times$ & 62.52 & 2.16$\times$\\
\SCD\ (Left and right context) & 91.20 & 1.15$\times$ & 92.57 & 0.85$\times$ & 92.11 & 1.13$\times$ & 85.46 & 0.63$\times$\\
\midrule
& \multicolumn{8}{c}{Spider} \\
\midrule
Unconstrained & 70.50 & 1.00$\times$ & 80.08 & 1.00$\times$ & 73.69 & 1.00$\times$ & 65.38 & 1.00$\times$\\
\sysname & 98.94 & 0.32$\times$ & 96.13 & 0.29$\times$ & 99.23 & 0.43$\times$ & 99.03 & 0.16$\times$\\
\sysname\ (No Two-phase decoding)  & 99.03 & 2.30$\times$ & 96.13 & 1.28$\times$ & 99.23 & 2.83$\times$ & 99.03 & 1.22$\times$\\
\SCD\ (Left and right context) & 99.13 & 0.33$\times$ & 99.13 & 0.29$\times$ & 99.13 & 0.44$\times$ & 99.03 & 0.16$\times$\\
\bottomrule
\end{tabular}
\end{small}
\end{table*}

\begin{table}[h]
\caption{Robustness of template grammars compared to Baseline on BIRD dev set. \SCD\ is followed for all these methods where Gold SQL in modified in different ways such as converting to small case, replacing single whitespace with random whitespaces, pretty-printing the template grammar. Significant drop is observed in ExM for all models except Granite.}
\label{tab:robustness-study}
\begin{tabular}{lrrrrrrrr}
\toprule
 Gold SQL Modification & CodeS-1B & CodeS-15B & Llama-3.1-8B & Granite-3.1-2B\\
\midrule 
\SCD\  & 91.20 & 92.57 & 92.11 & 85.46 \\
Small Case SQL & 90.61 & 92.44 & 90.81 & 86.83 \\
Pretty Format & 91.40 & 92.76 & 88.92 & 86.31 \\
Random Spaces (2,3) & 34.88 & 69.23 & 79.99 & 88.14 \\
Random Spaces (2,5) & 32.07 & 64.99 & 57.50 & 83.70 \\
\bottomrule
\end{tabular}
\end{table}

\subsection{Template Constrained Decoding}
\label{sec:temp_cons_decoding}
Grammar-constrained decoding is essential for ensuring the syntactic validity of SQL queries generated by LLMs. However, existing methods often suffer from high latency and are sensitive to formatting variations. In this section, we evaluate \sysname, our proposed method that aims to balance correctness and efficiency.  We benchmark all methods on the BIRD and Spider dev sets, which contain 1534 and 1034 queries respectively.  Table~\ref{tab:efficient-docoding-results} presents performance comparisons across different decoding strategies:

\begin{enumerate}[leftmargin=0.3cm, itemsep=0pt]
\item \textbf{Unconstrained:} This is the standard decoding setup where the SQL is generated without any template constraints.    
As expected, unconstrained generation is generally faster than grammar constrained methods.  However, this comes at the cost of significantly lower execution accuracy (ExM). 
\item \textbf{\sysname}:  We next compare the constrained generation algorithm of \sysname\ where we use the flexible grammar with the efficient two-phase partitioning decoding algorithm~\ref{alg:efficient_decoding}.  We observe huge jump in accuracy with constrained decoding without incurring latency overheads.
On the BIRD dataset, \sysname\ achieves latency comparable to — and for some LLMs (CodeS15B and Granite), even better than — unconstrained decoding. On Spider, latency improvements range from $2$ - $6\times$ over the unconstrained method. \sysname's accuracy and efficiency gains are due to important design decisions, and we present an ablation on each of these next.
\item \textbf{\sysname\ Without Two-phase Decode}:  If we run the flexible grammar with the default GCD algorithm~\ref{alg_fixed_g} instead of the two phase efficient inference of algorithm~\ref{alg:efficient_decoding}, we incur a factor of two to three times latency overhead compared to unconstrained generation.  This shows that off-the-shelf GCD methods are expensive.   The slight accuracy drop arises because \sysname\ supplies the entire partition in a single generate call, which alters the logit distribution due to layer normalization effects.
\item \textbf{\sysname\ without Context Tokens}: Another crucial factor for accurate generation with partitioned decoding was to include left and right context tokens to adjust for LLM's tokenization that can straddle across partition boundaries.  We oberve that dropping the context tokens causes huge drop in accuracy.  On the BIRD dataset, accuracy drops from above 90\% to around 80\% when averaged across LLMs.  For some LLMs, example Llama the drop is huge --- from 92\% to 79\%.  We present some anecdotes:
    \begin{lstlisting}[mathescape]
Examples:
1. SELECT T2.`School Name` ... `Enrollment (K-12)` > 0.1 AND NumGE1500 > 0
2. SELECT Website FROM schools ... AdmLName1 = 'Larson') OR (AdmFName1 = 'Dante' AND AdmLName1 = 'Alvarez')
    \end{lstlisting}
    In example 1, if the next token " AND" is not provided as part of the grammar when generating the literal 0.1, the LLM may repeatedly generate numbers instead of predicting an <eos> token. Similarly, in example 2, omitting the \textquotesingle)\textquotesingle\ token from the grammar while generating \textquotesingle{Larson}\textquotesingle could lead to malformed outputs. This is because missing the next token ID results in over-masking of valid token completions.
\item \textbf{\SCD:} Next we establish the usefulness of the flexible grammar by replacing with the Fixed template, an example of which is shown in the second row of Table~\ref{tab:templates_snippets}.  We report accuracy of this grammar too both with and without the context tokens.  We observe that without the context tokens, the fixed template method shows much bigger accuracy drops. When we extend the fixed grammar with left and right context tokens, the accuracy does bounce back to be almost comparable to what we obtained with the flexible grammar.   
\end{enumerate}
    Across multiple models and datasets (BIRD and Spider), the two-phase decoding approach shows $1.5\times$ to $2.3\times$ improvements in latency over the vanilla constrained decoding method (referring to rows “\sysname” and “\sysname (No Two-phase decoding)” in Table \ref{tab:efficient-docoding-results} for both BIRD and Spider. These results demonstrate that our method reduces unnecessary computation and improves efficiency during inference.
    Overall, it may appear that \GCD\ does not provide much gains beyond \SCD.  That may be because current LLMs are already exposed to the SQL formatting deployed in public Text-to-SQL benchmarks like BIRD or Spider.  We show that the accuracy of \SCD\ is highly sensitive to the surface formatting of the SQL template.
    Table~\ref{tab:robustness-study} shows results for various perturbations: converting only SQL keywords to lowercase (literal values remain case-sensitive), randomly replacing single spaces between keywords with 2–5 or 2–3 spaces, and Pretty-Format where we reformat the SQL using standard indentation and line breaks using SQLGlot to improve readability. 
    CodeS and LlaMA models show sharp accuracy drops with these formatting changes, whereas Granite remains more robust but the best accuracy of the Granite model is lower. These findings reinforce that fixed templates created out of the given SQL are highly sensitive to formatting of $\ty$.  The accuracy of \sysname\ is invariant to the string form of the template because it expresses all templates using a flexible grammar (shown in the last row of Table~\ref{tab:templates_snippets}).

\section{Conclusion and Future Work}
\label{sec:conclusion}
\paragraph{Conclusion} We introduced \sysname, a system designed to significantly boost Text-to-SQL accuracy by utilising templates derived from frequently occurring queries. \sysname\ converts labelled Text-to-SQL pairs into reusable templates and employs an accurate matching model to identify user queries that conform to these templates. For matched queries, \sysname\ uses a template-constrained decoding process, achieving substantial accuracy improvements (up to 36\% over ICL) and enhanced inference latency (1.5-2.2x faster). 

Most significantly, the paper establishes that for enterprise workloads with recurring query patterns, which can represent more than 50\% of queries in real-world settings, \sysname\ provides a practical path to high-accuracy Text-to-SQL conversion. This approach is particularly valuable as it works without fine-tuning and its design for incremental deployment, making it particularly suitable for enterprise environments where workloads evolve. The system's accuracy and coverage are not static; as new, expert-verified NLQ-SQL pairs become available, they can be seamlessly converted into templates and added to the template pool, allowing the system to continuously adapt and improve its performance. Crucially, TeCoD provides a robust fallback mechanism. When no matching template is found, the system defaults to a standard generation method using in-context examples, ensuring that it can service both frequent, recurring queries and novel, ad-hoc ones.

The extensive evaluations on both the BIRD and Spider benchmarks across different LLMs consistently demonstrate that \sysname\ outperforms both zero-shot and in-context learning approaches, often by substantial margins. The error analysis provides crucial insights into remaining challenges as to why  \sysname\ generated incorrect SQL in a small number of cases.

\paragraph{Future Work}
The current implementation focuses on templates where only literals are masked. Future research could explore more flexible template representations. While \sysname\ excels at handling head queries, tail queries still rely on standard in-context learning. Exploring hybrid approaches that combine elements of template-based generation with more flexible generative techniques could help improve performance across the entire query distribution. The principles of identifying recurring patterns and enforcing structural constraints during generation are likely to be broadly applicable.

\begin{acks}
We acknowledge the support of the SBI Foundation Hub for Data Science \& Analytics at the Indian Institute of Technology Bombay for providing financial support and infrastructure for conducting the research presented in this paper.
\end{acks}

\appendix
\section{Appendix}
\subsection{Analysis of NL similarity and template match success}
\label{app:nl_similarity}
\begin{figure}[h]
    \centering
    \includegraphics[width=1\columnwidth]{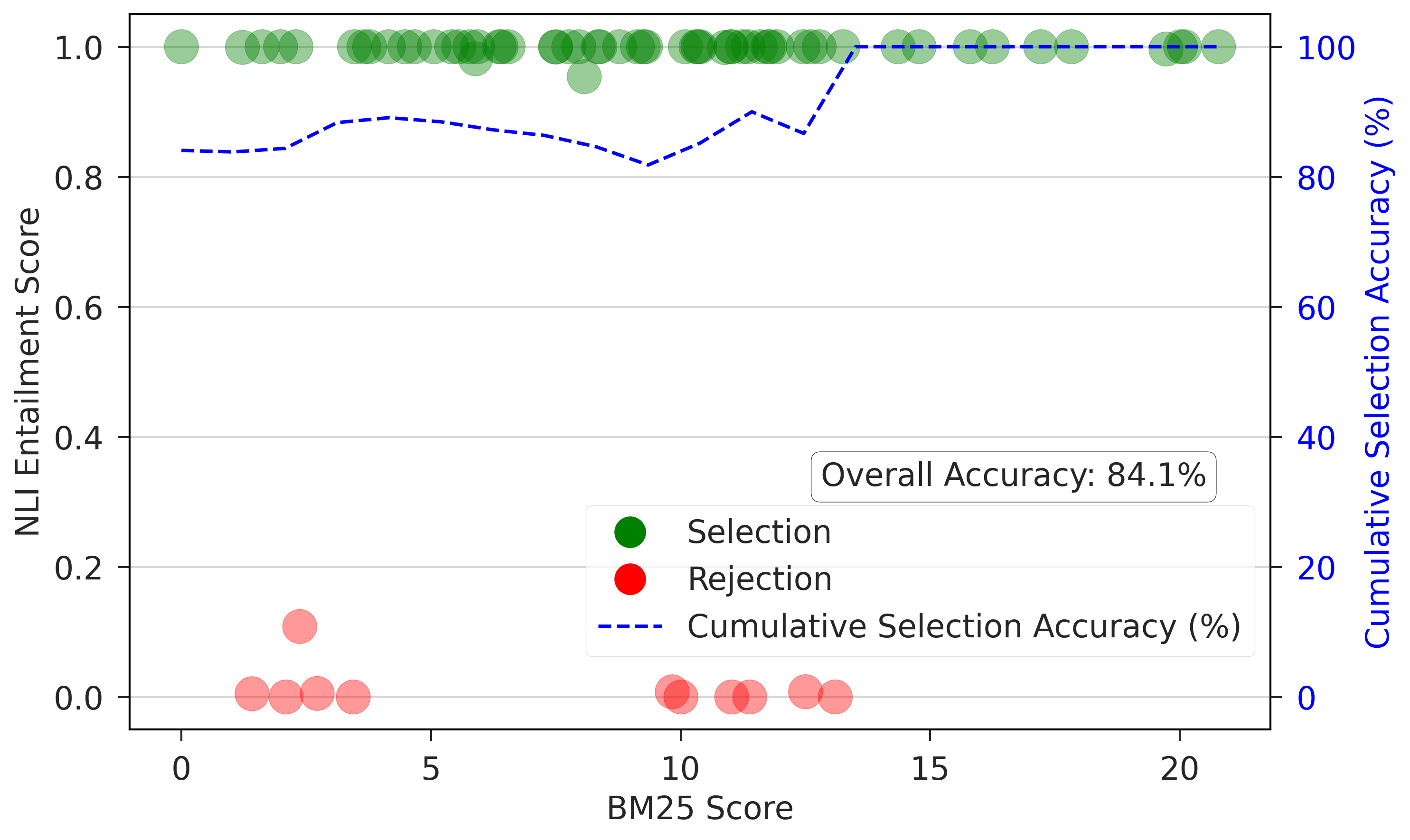}
    \caption{Scatter plot of BM25 score of the query with its most similar alternate NLQ from the matching template over Bird-Non-synthesized matched test set.}
    \label{fig:simMatch}
\end{figure}

We analyze how similar a query has to be with a stored template for the template matcher to work.  For this, in Figure~\ref{fig:simMatch} we present a scatter plot of the BM25 similarity of the query with its template against the match probability output by the template matcher. We observe that cumulative selection accuracy increases as expected as BM25 similarity increases, but even at low similarity levels, we get accuracy close to 80\%.  also present a few examples where the template matcher selected and rejected the template in Table~\ref{tab:template_examples}.

\begin{table*}[htbp]
\centering
\caption{Examples of template matcher performance, showing cases of successful and failed matches based on BM25 similarity.}
\label{tab:template_examples}

\begin{tabular}{l|c|p{0.325\textwidth}|p{0.325\textwidth}}
\toprule
\textbf{Outcome} & \textbf{BM25 Score} & \textbf{Query} & \textbf{Template} \\
\midrule
\multicolumn{4}{c}{\textit{Rejection Cases}} \\
\midrule
Rejection & 2.10 & Please list the leagues from Germany. & Tell me the name of the country's football league for \_. \\
Rejection & 2.37 & Give the number of "Revival" badges. & What is the tally of users who obtained the \_ badge? \\
\midrule
\multicolumn{4}{c}{\textit{Selection Cases}} \\
\midrule
Selection & 1.22 & In the non-carcinogenic molecules, how many contain chlorine atoms? & Retrieve the total number of molecules known to cause \_ that have \_ in their makeup. \\
Selection & 1.62 & What is Abomination's superpower? & Give me the super abilities that \_ is known for. \\
\bottomrule
\end{tabular}
\end{table*}

\begin{table}[h]
\centering
\caption{Dataset Statistics}
\label{tab:dataset_stats}
\setlength{\tabcolsep}{3.5pt} 
\begin{tabular}{lrrrr}
\toprule
\textbf{Dataset} & \textbf{\#Databases} & \textbf{Matched} & \textbf{Unmatched} & \textbf{Template} \\
\midrule
BIRD-Syn & 11 & 0 & 764 & 770 \\
Spider-Syn & 20 & 0 & 525 & 509 \\
BIRD-Real & 10 & 73 & 61 & 73 \\
Spider-Real & 16 & 484 & 15 & 482 \\
\bottomrule
\end{tabular}
\end{table}

\subsection{Analysis of literal generation errors}
\label{app:error}
\begin{enumerate}
    \item Templates using a particular SQL syntax were consistently associated with incorrect literal generation. Since all queries with this syntax resulted in errors, it is plausible that the LLM had limited or no exposure to such patterns in its training data.
    \begin{lstlisting}[mathescape, escapeinside=||]
Example:
Gold: SELECT CAST(`Free Meal Count (K-12)` AS REAL) / `Enrollment (K-12)` FROM frpm ORDER BY `Enrollment (K-12)` DESC LIMIT 9, 2
Pred: SELECT cast( "Free Meal Count (K-12)" AS REAL ) / "Enrollment (K-12)" FROM frpm ORDER BY "Enrollment (K-12)" DESC LIMIT |\textcolor{red}{10}| OFFSET 9        
    \end{lstlisting}

    \item In some cases, the schema item and its correct literal value are included in the prompt, yet the LLM generates a slightly different version of the literal, such as \texttt{\textquotesingle=\textquotesingle} being produced as \texttt{\textquotesingle{ = }\textquotesingle}. In such cases often disagreement is observed between schema item value presented by schema and evidence in prompt. 
    \begin{lstlisting}[mathescape, escapeinside=||]
Example:
Prompt: database schema :
table bond , columns = [ bond.bond_type ( text || values : - , = ) , bond.molecule_id ( text || values : TR000 , TR001 ) , bond.bond_id ( text || primary key || values : TR000_1_2 , TR000_2_3 ) ] ...
Evidence:double bond refers to bond_type = ' = '
Question: Please list top five molecules that have double bonds in alphabetical order.
Gold:SELECT DISTINCT T.molecule_id FROM bond AS T WHERE T.bond_type = '=' ORDER BY T.molecule_id LIMIT 5
Pred:SELECT DISTINCT T.molecule_id FROM bond AS T WHERE T.bond_type = |\textcolor{red}{\texttt{\textquotesingle{ = }\textquotesingle}}| ORDER BY T.molecule_id LIMIT 5
    \end{lstlisting}
    A possible explanation is the recency bias often observed in LLMs, since the evidence snippet (with extra whitespace) appears closer to the end of the prompt.

    \item When generating format specifiers for dates, the LLM often produces incorrect literals.
    \begin{lstlisting}[mathescape, escapeinside=||]
Example:
Gold: SELECT DISTINCT T1.ID, STRFTIME('%Y', CURRENT_TIMESTAMP) - STRFTIME('%Y', T1.Birthday) FROM Patient AS T1 INNER JOIN Examination AS T2 ON T1.ID = T2.ID WHERE T2.RVVT = '+'
Pred: SELECT DISTINCT T1.ID, strftime(|\textcolor{red}{\texttt{\textquotesingle\%J\textquotesingle}}|, CURRENT_TIMESTAMP) - strftime(|\textcolor{red}{\texttt{\textquotesingle\%J\textquotesingle}}|, T1.Birthday) FROM Patient AS T1 INNER JOIN Examination AS T2 ON T1.ID = T2.ID WHERE T2.RVVT = '+'
    \end{lstlisting}

    \item On the Spider dataset, CodeS-15B-Spider shows a drop in Execution Match (ExM) due to consistent errors in string literal generation. The model overfits to formatting with two spaces between the comparison operator and the string literal, failing when only one space is allowed likely due to exposure to such formatting in the training data. When evaluation permits variable spacing, it generates correct literals, indicating the issue stems from overfitting to surface formatting rather than misunderstanding the query.
    \begin{lstlisting}
    
    \end{lstlisting}
\end{enumerate}

\subsection{Embedding and NLI model}
\label{app:nli_model}
As detailed in the paper, our template matching module initially utilized the NV-Embed-v2 model for creating embeddings and a fine-tuned distilbert-base-uncased model for the Natural Language Inference (NLI). While evaluating for the non-synthetic workload, we have updated these components to use more recent, state-of-the-art models. Specifically, we now employ Qwen/Qwen3-Embedding-4B \cite{qwen3embedding} to generate embeddings and a fine-tuned Qwen/Qwen3-Reranker-4B \cite{qwen3embedding} for the NLI classification task. The core methodology is unchanged: the embedding model is used directly to perform an initial similarity search for candidate retrieval, and the reranker is subsequently fine-tuned and used as the NLI model to validate and select the final template. However, we found using the mean of NLI scores of alternates from the same template to be better and used the same for selecting the template.


\bibliographystyle{ACM-Reference-Format}
\bibliography{parsing,custom,pubs}

\end{document}